# A Multi-agent *Large Language Model Framework to Automatically Assess Performance of a Clinical AI Triage Tool*

Adam E. Flanders, MD[1]; Yifan Peng, PhD[2]; Luciano Prevedello, MD MPH[3]; Robyn Ball, PhD[4]; Errol Colak, MD[5]; Prahlad Menon, PhD[1]; George Shih, MD[2]; Hui-Ming Lin, BS[5]; Paras Lakhani, MD[1]

[1]Thomas Jefferson University, Philadelphia, Pennsylvania
[2]Weil-Cornell Medical Center, New York, New York
[3]The Ohio State University, Columbus, Ohio
[4]The Jackson Lab, Bar Harbor, Maine
[5]Unity Health, Toronto, Canada

Corresponding Author: Adam E. Flanders, MD, Thomas Jefferson University, Suite 1080B Main Building, 132 S. Tenth Street, Philadelphia, PA 19107, USA; Voice (215) 955-2430; Fax (215) 955-5329; Email adam.flanders@jefferson.edu

**Title:** *A Multi-agent Large Language Model Framework to Automatically Assess Performance of a Clinical AI Triage Tool*

**ABBREVIATIONS**

LLM = large language model, DICOM = digital imaging and communication in medicine, ICH = intracranial hemorrhage, AP = average precision, AUC = area under the curve, MCC = Matthews Correlation Coefficient

**KEY RESULTS**

- An ensemble of nine LLM agents analyzed report impressions for ICH from 29,766 head CTs. The highest AUC performance was with llama3.3:70b and GPT-4o (AUC= 0.78).
- The average precision was highest for Llama3.3:70b and GPT-4o (AP=0.75 & 0.76). Llama3.3:70b had the highest F1 score (0.81), recall (0.85), precision (0.78), specificity (0.72) and MCC (0.57).
- The ideal ensemble of LLM agents using MCC (95% CI) values were: Full-9 Ensemble 0.571, Top-3 Ensemble 0.558, Consensus 0.556, and GPT4o 0.522.

**SUMMARY STATEMENT**

- A consensus reached by a local open-source ensemble of LLM agents offers a more reliable solution to monitor performance of radiology AI triage tools over a single LLM when using the radiology report as the reference standard.

**ABSTRACT**

**Purpose:** Large language models (LLM) have the capability to abstract concepts from free text in radiology reports. LLMs vary in performance in extracting specific concepts. The purpose of this study was to determine if an ensemble of multiple LLM agents could be used collectively to provide a more reliable assessment of a pixel-based AI triage tool.

**Methods:** 29,766 non-contrast CT head exams from fourteen hospitals were processed by a commercial intracranial hemorrhage AI detection tool. The impression from the radiology reports was processed by an ensemble of eight open-source LLM models and a HIPAA compliant internal version of GPT-4o using a single multi-shot prompt that assessed for presence and type of intracranial hemorrhage. 1,726 examples were manually reviewed. Performance characteristics of the eight open-source models and consensus were compared to GPT-4o. Three ideal consensus LLM ensembles were tested for rating the performance of the triage tool.

**Results:** Cohort consisted of 29,766 head CTs, 15,189 female, 14,564 male, mean age 62.76 ± 20.09. The highest AUC performance was achieved with llama3.3:70b and GPT-4o (AUC= 0.78). The average precision was lowest for Llama3.2:1b (AP=0.55), while highest for Llama3.3:70b and GPT-4o (AP=0.75 & 0.76). Llama3.3:70b had the highest F1 score (0.81) and recall (0.85), greater precision (0.78), specificity (0.72), and MCC (0.57). In testing the ideal combination of LLMs to use to assess performance of the ICH detector, MCC (95% CI) values were: Full-9 Ensemble 0.571 (0.552–0.591), Top-3 Ensemble 0.558 (0.537–0.579), Consensus 0.556 (0.539–0.574), and GPT4o

0.522 (0.500–0.543). No statistically significant differences were observed between Top-3, Full-9, and Consensus ($p > 0.05$).

**Conclusion:** An ensemble of medium to large sized open-source LLMs provides a more consistent and reliable method to derive a ground truth retrospective evaluation of a clinical AI triage tool over a single LLM alone.

**INTRODUCTION**

Many users of clinical artificial intelligence (AI) tools have little objective evidence to determine if the tools are regularly utilized or whether the tools effectively augment clinical care. Not all vendors monitor local performance of their models, and even when they do, it is invariably incomplete. Ongoing monitoring of commercial AI tools is crucial to detect potential drift, as studies indicate that over 90% of AI models can experience temporal performance degradation [1,2]. The etiology of AI performance drift is multi-factorial, but it is most often attributed to subtle changes in the input data (e.g., new protocols, image processing, device addition or replacement). Alterations in the local patient population and disease prevalence can also contribute to the performance drift of an AI tool [3,4].

The resources required to assess real-time performance of an AI tool are not trivial, especially when establishing a reference standard requires reviewing electronic medical record (EMR) data, radiology reports, or the original imaging [5,6]. The time and expertise required to manually review data retrospectively to obtain ground truth verification of an AI prediction are prohibitive. Requesting radiologists to verify AI results in real-time is often met with resistance, and compliance can be poor. As an alternative the final radiology report has served as a surrogate for ground truth verification. Natural language processing (NLP) techniques have been employed as a means to avoid human expert review of reports; however, they depend heavily on encoding all possible scenarios, making them less effective to variations in reporting style [7]. Large language models (LLMs) have been shown to successfully extract and normalize concepts from heterogeneous free text in radiology reports although performance varies with the model used and the task presented [8–10]. A number of LLMs are freely available in the open-source community and can be locally deployed to mitigate patient privacy concerns related to the exposure of protected health information (PHI) to outside commercial entities. An ensemble of LLMs with a majority consensus has been shown to exhibit superior performance over a single LLM for certain tasks [11].

Because of the variability in observed individual LLM performance, the purpose of this study was to determine if an *ensemble* of open source LLMs instead of a single LLM provides a more reliable and consistent method to obtain ground truth. This methodology is evaluated as a practical application to automate measuring the performance of a commercial intracranial hemorrhage (ICH) AI triage detection tool.

**MATERIALS & METHODS**

Approval was obtained by the local institutional review board for this HIPAA compliant retrospective review of data. The data collected focused on a prospective collection of results from an ICH detector developed by Viz.AI (San Francisco, CA). Inference was performed on consecutive emergency room and outpatient non-contrast head CTs from fourteen hospitals which included two community practices supported by teleradiology and an academic core. DICOM routing to the AI inference engine was predominately focused on emergency room exams, however some inpatient and outpatient exams were included in the cohort [Table 1 and Figure 1]. All AI results were made available to the radiologist at the time of dictation. The radiologists did not provide any direct feedback on the AI results. The automated system that assembled the data for this study (Real-time AI Data Assessment and Reporting or RADAR) was constructed using off-the-shelf components [Figure 2].

The impression text was parsed from the report and combined with a set of multi-shot prompt instructions (Supplement A) requesting a specific response if the concept of *acute intracranial hemorrhage* was present or absent in the impression text. The prompt specifies that a JSON object be returned with the presence of (Boolean) and if positive, the type of hemorrhage. If positive, the key-value pair is provided as `[hemorrhage : <true / false>, subtype: <subarachnoid, intraparenchymal, subdural, epidural, intraventricular hemorrhage>]`. Otherwise, the key-value pair is provided as `[hemorrhage : false]`. The combined prompt and impression text was processed by local versions of eight LLMs of various sizes and types (Llama3.2:1b, Llama3.2:3b, CodeLlama:7b,

Llama3.1:8b, Granite3-dense:2b, Llama3.3:70b, Granite3-dense:8b, and DeepSeek-r1) managed through the Ollama engine API (Meta, San Francisco, CA). Each LLM processed the same report in succession with the temperature hyperparameter set to zero. An additional consensus score was created based upon agreement of four or more models.

An internal, HIPAA-compliant Microsoft Azure (Redmond, WA) hosted version of GPT-4o (OpenAI, San Francisco, CA) was the ninth model evaluated on the same corpus of report impressions and multi-shot prompt as a separate exercise. The results of GPT-4o were matched and compared to each of the original eight LLMs and the consensus of eight.

In order to assess the level of agreement among the the LLMs in the identification of the hemorrhage concept in the impression text, Cohen's *kappa* and Jaccard similarity [12] were employed for each LLM paired against each other, and the LLM ensemble consensus vote was reached from an ensemble where four or more LLMs were in agreement.

A sample of the radiology report impressions was reviewed by two of the authors: a neuroradiologist with over 30 years of experience and a cardiothoracic radiologist with over 15 years of experience. All of the discordant result-report pairs, defined as disagreements between the LLM consensus and result from the ICH AI tool were scrutinized primarily to provide a conservative estimate of the LLM performance overall in addition to a small sample of the concordant exam-report pairs. Reports were categorized by the reviewers as absolute positive, absolute negative, incomplete information, or indeterminate (where the concept of hemorrhage was not discernible because of ambiguous descriptions or incomplete information (e.g. no change). Exams with ambiguous or incomplete report impressions for the concept of hemorrhage were excluded from the performance analysis.

Receiver-operator and precision-recall analysis were generated for all nine LLM agents to determine how consistently each LLM predicted the presence of hemorrhage compared to the human-reviewed reports.

Using the human-reviewed reports as the reference standard, performance characteristics were also calculated for each LLM, yielding recall, specificity, accuracy, precision, Cohen's kappa, MCC and F1 score. A composite performance score was derived as the arithmetic mean of the eight standard metrics (accuracy, precision, recall, specificity, negative predictive value, F1-score, Cohen's κ, and MCC), normalized by seven to balance weighting across positive and negative predictive performance.

Various LLM configurations were evaluated to identify an ideal combination to use as ground truth assessment tool in evaluating the performance of the ICH detection algorithm. This included a consensus of all eight LLM models where four or more were in agreement, a smaller ensemble using the top three performing LLM models, GPT-4o alone, and a consensus of the entire ensemble of nine models where five or more models were in agreement. Paired bootstrap comparison of the Matthews correlation coefficient (MMC) was performed for each scenario to determine if there was a significant difference in performance characteristics between each ensemble.

All of the manuscript text was generated by the authors. GPT-4o and GPT-5 (OpenAI, San Francisco, CA) was used to construct and verify some of the statistical python code.

**RESULTS**

*Dataset*

The data cohort is summarized in Table 1 consisting of 29,766 non-contrast CT head examinations performed between July 2024 and April 2025, derived from 14 hospitals in two-state health systems. DICOM routing gave preference to emergency studies (26,165), however, 1,986 inpatient and 615 outpatient head CTs were also included in the cohort. The cohort consisted of 14,564 males, 15,189 females and 12 of unknown gender. Age ranged from 15 to 100 with a mean age of 62.76 ± 20.09 (95% CI: 62.53, 62.99).

The Viz.ai ICH AI model detected ICH in 1,469 of 28,297 (4.93%) of exams. The number of exams that failed inference was not captured.

*Assessing concordance between LLMs*

Cohen's kappa [Figure 3a] and Jaccard similarity [Figure 3b] were calculated for each pair of LLMs and separately calculated between each LLM and the consensus vote (four or more in agreement). Overall, we found that Llama3.2:1b had the lowest agreement with the other models. The overall agreement exceeded 0.7 for many of the medium and large-sized models with between eight and seventy billion parameters. GPT-4o alone showed excellent agreement with the ensemble consensus and DeepSeek-r1 alone.

*Comparing performance of LLM to human reviewed reports*

A total of 1,726 report impressions were manually reviewed for the presence/absence of acute ICH. Of these reports, 236 (14%) were judged to be either incomplete or ambiguous to make a valid determination. The remaining 1,490 that were reviewed were definitive as either positive or negative for ICH and served as a conservative reference standard for assessing LLM performance. Performance characteristics and composite score of the nine large language models and consensus (four or more of eight in agreement) to identify the concept of hemorrhage in the impression of the report was compared to human review (n=1,490) [Table 2]. Llama3.2:1b was the lowest performing model, whereas llama3.3:70b had the highest F1 score (0.81) and recall (0.85). Llama3.3:70b also had greater precision (0.78), specificity (0.72), and MCC (0.57) compared to others. Deepseek-r1 was not a top-performer relative to other models.

Receiver-operator characteristics for each large language model and consensus in predicting ground-truth using a human observer review of 1,490 report impressions as reference [Table 3]. The lowest discriminating power was observed for Llama3.2:1b (AUC=0.50). The highest performance was achieved with Llama3.3:70b and GPT-4o (AUC= 0.78). Precision-Recall characteristics for the LLMs and consensus are also shown in Table 3. The average precision was lowest for Llama3.2:1b (AP=0.55) highest for Llama3.3:70b and GPT-4o (AP=0.75 & 0.76)

*Gauging change in apparent performance of the AI ICH tool by varying configuration of ensembles*

To simulate how ensemble configuration might introduce variability into the assessment of ICH AI tools' performance, we created four different configurations of LLM consensus as a reference standard and used each to evaluate the performance of the Viz.ai ICH model [Table 4]. The configurations tested included: (1) an ensemble of the top three performing LLM (Llama3.3:70b, GPT-4o and Granite3-dense:8b), (2) a consensus of all nine models (including GPT-4o, using five or greater agreement), (3) consensus agreement of the eight LLM agents (four of more) and (4) GPT-4o alone using the entire dataset (n=29,766). Using all 29,766 cases, we compared AI result performance against four alternate ground truth definitions (Top-3 Ensemble, Full-9 Ensemble, GPT4o, and Consensus) using MCC. Bootstrap resampling (1,000 replicates) was used to generate 95% confidence intervals and perform paired comparisons. MCC (95% CI) values were: Full-9 Ensemble 0.571 (0.552–0.591), Top-3 Ensemble 0.558 (0.537–0.579), Consensus 0.556 (0.539–0.574), and GPT4o 0.522 (0.500–0.543). Paired bootstrap testing indicated significantly lower MCC for GPT4o compared with Top-3 (p = 0.026), Full-9 (p < 0.001), and Consensus (p = 0.020), with no significant differences between Top-3, Full-9, and Consensus (p > 0.05).

**DISCUSSION**

The objective of this study was to determine whether an ensemble of LLM agents could be more effective than a single model in automatically monitoring the performance of a commercial ICH detection AI model. Under ideal circumstances, access to multiple *visual* models could serve to adjudicate the commercial model, but the cost for this type of solution is prohibitive. Use of the clinical report as a surrogate for ground truth is a reasonable alternative when combined with an LLM. It has been shown that an ensemble of LLM agents serves as a surrogate for multiple observers, providing a means for self-correction [11,13]. In this study, an ensemble of LLMs evaluated only the impression section of the report, with the expectation that a group of LLMs would make up for the deficiencies of any single LLM.

Our adjudication process focused specifically on examples where either a true disparity occurred between radiologist and ICH detection, or a failure of the LLM ensemble. It was reasoned that if neither human expert or an LLM could ascertain a reasonable level of certainty, they should be removed from analysis.

Rohren et al used a HIPAA complaint Microsoft Azure hosted version of GPT-4o to review radiologist reports from 332,809 head CT exams from 37 radiology practices and compared the LLM derived results as a means to automate post-deployment performance of a commercial ICH detector [14]. A total of 13,569 (4%) exams were labeled positive for ICH by the ICH detection model. Two hundred of the positive exams were randomly selected for manual review of both images and reports to derive performance metrics for the ICH detector, LLM, and radiologists. They found a high diagnostic accuracy in identifying the correct concept of ICH in the reports with only a single false negative and no ambiguous report content.

The decision to use open-source models was intentionally aimed at determining if the solution could be operationalized internally without the need to employ a HIPAA-compliant commercial version of an LLM. We compared the performance of our open-source ensemble solution to the HIPAA-compliant version of GPT-4o and found that their performance was comparable or superior. The task was simplified by focusing on the impression section content alone.

Nearly fourteen-percent of the exams with discordant results included reports with ambiguous findings on secondary review. Ambiguous language has not been consistently addressed in the literature. Lack of clarity can exist both with the images themselves and the reports depending upon image quality, artifacts, surgical changes, non-acute fluid collections and unrelated findings (e.g. mineralization). Despite the relative simplicity of the task (i.e. hemorrhage present or absent in the impression) there was substantial variability in performance across the collection of LLMs. As expected, the larger models with the most parameters Llama3.1:8b (8 billion parameters), Llama3.3:70b (70 billion parameters ), and GPT-4o (1.76 trillion parameters) provided the most consistent results, whereas some of the smaller models Llama3.2:1b and CodeLlama:7b created more disparate results from consensus or the larger models. Moderate performance was realized with smaller models Llama3.2:3b and Granite3-dense:2b suggesting that model design was a performance factor.

Prompt construction is a major contributor to LLM performance [8,10,15–17]. For example, our study employed a few-shot prompt, providing examples for the five subtypes of hemorrhage under investigation. The prompt only specified "intracranial" to define the anatomic region specifically and as a result swelling or hematomas on the scalp were often misclassified as intracranial. This underscores the requirement to perform testing and continuous monitoring of the LLM to identify patterns where the majority of the LLMs fail. Rohren et. al designed their LLM ICH detection prompt using a corpus of reports prior to deployment [14]. The use of the ensemble method does provide some capacity to correct for certain types of errors that were more prevalent with a single model and can correct for random misclassifications and biases yielding a more stable result by consensus similar to any multi-observer study. Interestingly, the larger models were not consistently more reliable than some of the medium sized models.

Limitations of this study include the nature of the dataset. ICH has a low prevalence in the general population and this is reflected by the large class imbalance in our data (4.7%) which is somewhat lower than the general estimates in the emergency department population (6.7%) [18]. This inflates specificity and accuracy values. There is an element of incorporation bias as the interpreting radiologist had access to the AI outputs. An evaluation-set bias is also present because we focused on discordant exams primarily; however, this provided a conservative estimate of the LLM performance rather than artificially enhancing performance. Dropping the ambiguous impressions from the analysis also introduced some bias in LLM performance.

**CONCLUSION**

A consensus evaluation generated by an open-source HIPAA compliant LLM ensemble provides a comparable or superior performance to commercial solutions and reliable automation of performance of clinical triage AI tools is achievable using available general purpose LLMs.

**TABLES**

| Exam type: | Non-contrast head CT |
|---|---|
| AI Feature being assessed: | Presence/absence of intracranial hemorrhage |
| Total: | 29,766 |
| **Gender** | |
| Male: | 14,564 |
| Female: | 15,189 |
| Unknown: | 12 |
| **Age** | |
| Mean ± Std Dev | 62.76 ± 20.09 |
| Range (years) | 14.9 – 102 yrs |
| **Setting** | |
| Emergency: | 26,165 |
| Inpatient: | 2,986 |
| Outpatient: | 615 |

Table 1. Demographics of the final head CT dataset that underwent inference using a commercial grade intracranial hemorrhage computer vision tool.

| LLM | Accuracy | Precision (PPV) | Recall (Sensitivity) | Specificity | NPV | F1-score | Cohen's Kappa | MCC | Composite Score |
|---|---|---|---|---|---|---|---|---|---|
| **Llama3.2:1b** | 0.51 (0.48-0.53) | 0.55 (0.51-0.58) | 0.55 (0.51-0.58) | 0.45 (0.42-0.49) | 0.46 (0.42-0.49) | 0.55 (0.52-0.57) | 0.00 (-0.05-0.05) | 0.00 (-0.05-0.05) | 0.44 (0.41-0.47) |
| **Llama3.2:3b** | 0.69 (0.67-0.72) | 0.70 (0.67-0.73) | 0.75 (0.72-0.78) | 0.62 (0.58-0.66) | 0.67 (0.64-0.71) | 0.72 (0.70-0.75) | 0.37 (0.32-0.42) | 0.37 (0.32-0.42) | 0.65 (0.62-0.68) |
| **CodeLlama:7b** | 0.66 (0.63-0.68) | 0.74 (0.71-0.78) | 0.57 (0.53-0.60) | 0.77 (0.74-0.80) | 0.60 (0.56-0.63) | 0.64 (0.61-0.67) | 0.32 (0.28-0.37) | 0.34 (0.29-0.39) | 0.61 (0.59-0.64) |
| **Llama3.1:8b** | 0.69 (0.67-0.72) | 0.65 (0.63-0.68) | **0.95** (0.94-0.97) | 0.38 (0.34-0.42) | **0.87** (0.83-0.90) | 0.77 (0.75-0.79) | 0.35 (0.31-0.40) | 0.42 (0.38-0.46) | 0.67 (0.65-0.69) |
| **Granite3-dense:2b** | 0.67 (0.64-0.69) | 0.65 (0.62-0.68) | 0.86 (0.83-0.88) | 0.44 (0.41-0.48) | 0.72 (0.68-0.77) | 0.74 (0.71-0.76) | 0.31 (0.27-0.36) | 0.33 (0.29-0.39) | 0.63 (0.60-0.65) |
| **Llama3.3:70b** | **0.79** (0.77-0.81) | 0.78 (0.75-0.81) | 0.85 (0.82-0.87) | 0.72 (0.69-0.75) | 0.80 (0.76-0.83) | **0.81** (0.79-0.83) | **0.57** (0.53-0.61) | **0.57** (0.53-0.61) | **0.76** (0.74-0.78) |
| **Granite3-dense:8b** | 0.72 (0.69-0.74) | 0.75 (0.72-0.78) | 0.72 (0.69-0.75) | 0.71 (0.67-0.74) | 0.68 (0.64-0.71) | 0.73 (0.71-0.76) | 0.43 (0.38-0.47) | 0.43 (0.38-0.47) | 0.68 (0.65-0.70) |
| **DeepSeek-r1** | 0.67 (0.64-0.69) | 0.65 (0.62-0.68) | 0.83 (0.81-0.86) | 0.47 (0.43-0.50) | 0.70 (0.66-0.74) | 0.73 (0.71-0.75) | 0.31 (0.26-0.35) | 0.32 (0.28-0.37) | 0.62 (0.60-0.65) |
| **GPT-4o** | 0.78 (0.75-0.80) | **0.83** (0.80-0.85) | 0.75 (0.72-0.78) | **0.81** (0.78-0.84) | 0.73 (0.70-0.76) | 0.79 (0.76-0.81) | 0.56 (0.51-0.60) | 0.56 (0.51-0.60) | 0.75 (0.72-0.77) |
| **Consensus** | 0.65 (0.62-0.67) | 0.62 (0.59-0.65) | 0.88 (0.86-0.90) | 0.37 (0.33-0.40) | 0.72 (0.67-0.77) | 0.73 (0.71-0.75) | 0.26 (0.21-0.30) | 0.29 (0.24-0.34) | 0.60 (0.58-0.63) |

Table 2. Performance characteristics with confidence intervals (95% lower-upper) of the nine large language models and consensus (four or more of eight in agreement) to identify the concept of hemorrhage in the impression of the report compared to human review (n=1490) including a calculated composite score. The highest performing model in each category is in **bold**. Llama3.2:1b was the lowest performing model, whereas Llama3.3:70b had the highest F1 score (0.81) and accuracy (0.79). An intermediate sized model (Llama3.1:8b) demonstrated the highest recall (0.95) and NPV (0.87). GPT-4o had the highest precision (0.83) and specificity (0.81) compared to others. DeepSeek-r1 was not a top-performer relative to other models.

| LLM | ROC AUC (95% CI) | PR AUC (95% CI) |
|---|---|---|
| Llama3.2:1b | 0.501 (0.476-0.526) | 0.546 (0.518-0.574) |
| Llama3.2:3b | 0.684 (0.661-0.707) | 0.663 (0.634-0.691) |
| CodeLlama:7b | 0.666 (0.641-0.689) | 0.657 (0.627-0.685) |
| Llama3.1:8b | 0.668 (0.648-0.689) | 0.645 (0.619-0.670) |
| Granite3-dense:2b | 0.651 (0.628-0.673) | 0.634 (0.605-0.663) |
| Llama3.3:70b | **0.782 (0.761-0.803)** | 0.746 (0.719-0.773) |
| Granite3-dense:8b | 0.715 (0.692-0.739) | 0.691 (0.663-0.718) |
| DeepSeek-r1 | 0.649 (0.625-0.671) | 0.633 (0.604-0.662) |
| GPT4o | 0.780 (0.759-0.802) | **0.756 (0.728-0.785)** |
| Consensus | 0.623 (0.602-0.645) | 0.615 (0.586-0.643) |

Table 3. Area under the curve for receiver-operator (ROC AUC) and precision-recall (PR AUC) for each large language model and consensus in predicting ground-truth using a human observer review of 1,490 report impressions as reference. The lowest discriminating power was realized for llama3.2:1b (AUC=0.50). The highest performance was achieved with llama3.3:70b and GPT-4o (AUC= 0.78). The average precision was lowest for llama3.2:1b (AP=0.55) highest for llama3.3:70b and GPT-4o (AP=0.75 & 0.76).

| Metric | Top-3 LLM Ensemble | Full-9 LLM Ensemble | GPT-4o | Consensus |
|---|---|---|---|---|
| Accuracy | 0.95 | 0.95 | 0.95 | 0.94 |
| Precision (PPV) | 0.62 | 0.71 | 0.59 | 0.74 |
| Recall (Sensitivity) | 0.53 | 0.50 | 0.51 | 0.46 |
| Specificity | 0.98 | 0.98 | 0.98 | 0.99 |
| F1 Score | 0.57 | 0.59 | 0.55 | 0.57 |
| Cohen's Kappa | 0.55 | 0.56 | 0.52 | 0.54 |
| MCC | 0.55 | 0.57 | 0.52 | 0.56 |

Table 4. Variation in calculated performance of the ICH detection model depending on ensemble configuration of LLMs used. An ensemble of the top three performing LLM (Llama3.3:70b, GPT-4o and granite3-dense:8b), a consensus of all nine models (using five or greater agreement), consensus agreement of the eight Ollama LLM (four of more) versus GPT-4o alone using the entire dataset (n=29,766). Paired bootstrap testing revealed significant differences in MCC for GPT-4o compared with Top-3 Ensemble (p = 0.026), Full-9 Ensemble (p < 0.001), and Consensus (p = 0.020). No statistically significant differences were observed between Top-3, Full-9, and Consensus (p > 0.05).

**FIGURES**

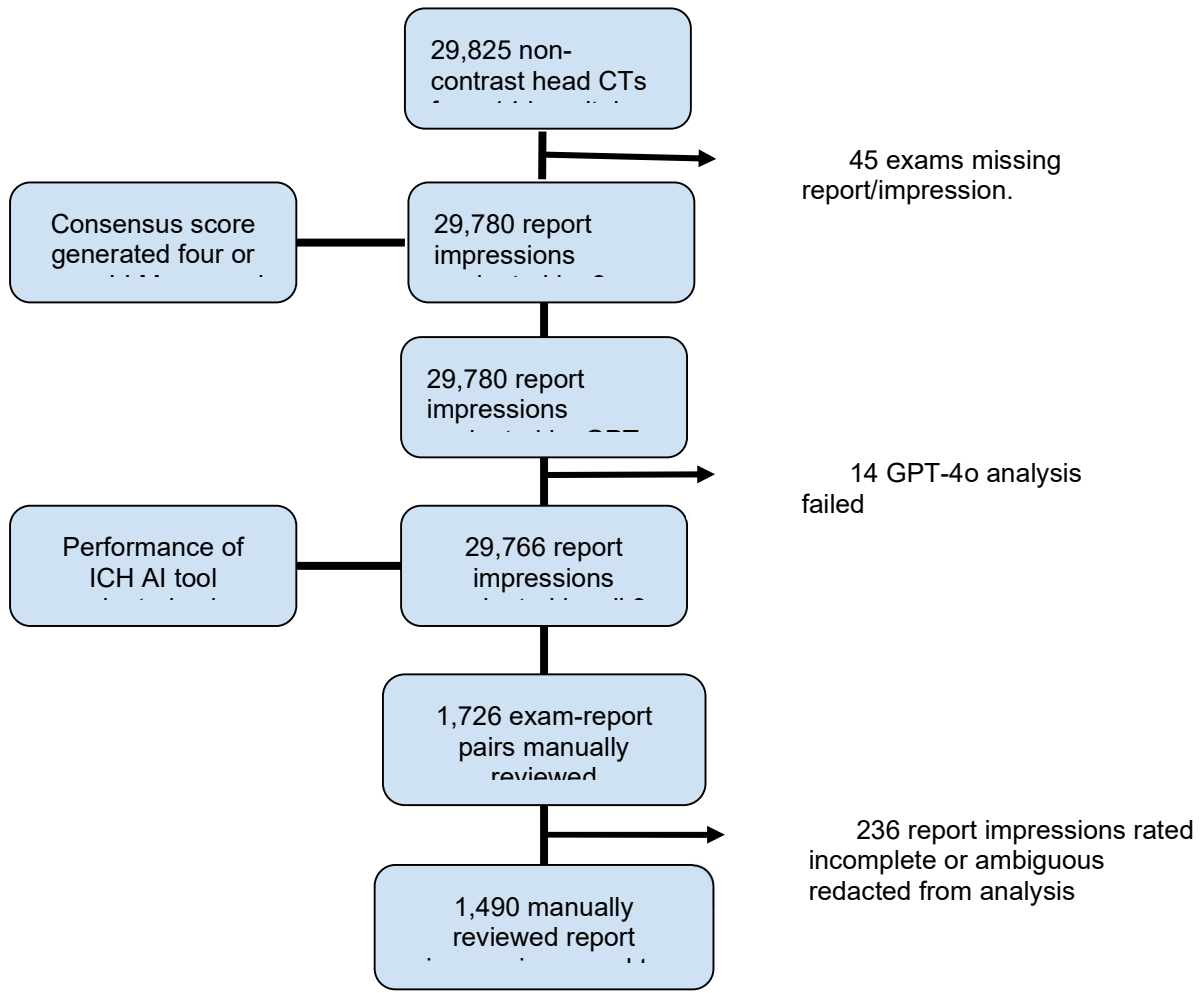

Figure 1. Data diagram. 29,825 non-contrast head CT exams run through Viz.AI ICH inference engine over a 14 month period. 45 exams had missing reports or impressions and 14 failed GPT-4o analysis and were removed from the cohort. 29,766 report impressions were rated by all nine LLM. 1,726 report impressions (6%) were manually graded and 236 were redacted because of ambiguous or incomplete information. 1,490 reviewed reports were used to assess performance of the nine LLMs.

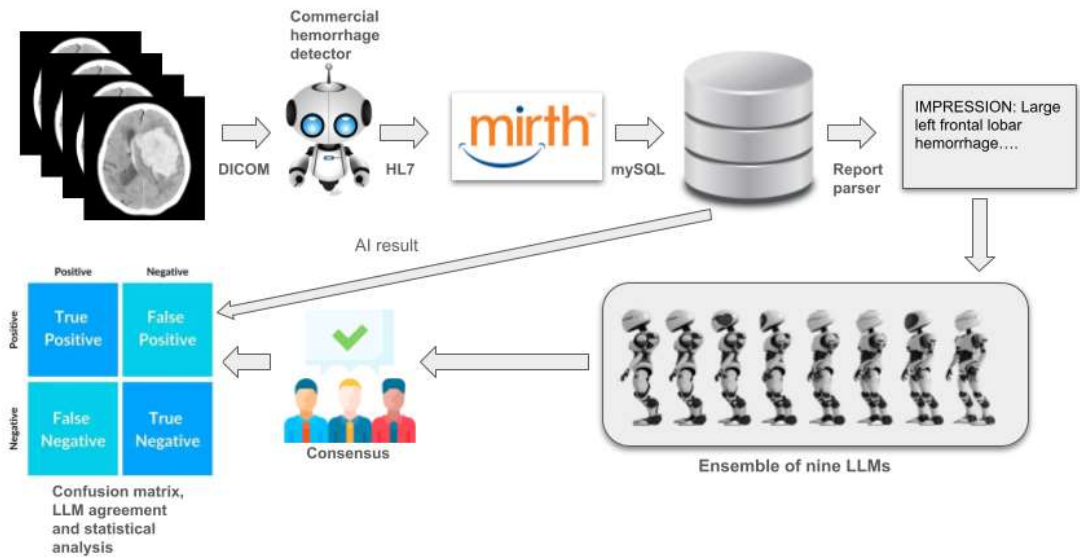

Figure 2. Overview of the Real-time AI Data Assessment and Reporting (RADAR) LLM automated performance engine. The vendor ICH detector receives the DICOM images for non-contrast head CT examinations and transmits a simple HL7 message with the boolean results for the presence or absence of hemorrhage. The HL7 result message is processed by a generic MIRTH receiver, which parses the message to add a row in a local MySQL database. Once a day, a script extracts reports from the reporting system using the Powerscribe API, it parses the impression from the report, and passes the impression, combined with a specific prompt, to each of eight LLMs. A consensus is generated when four or more models are in agreement.

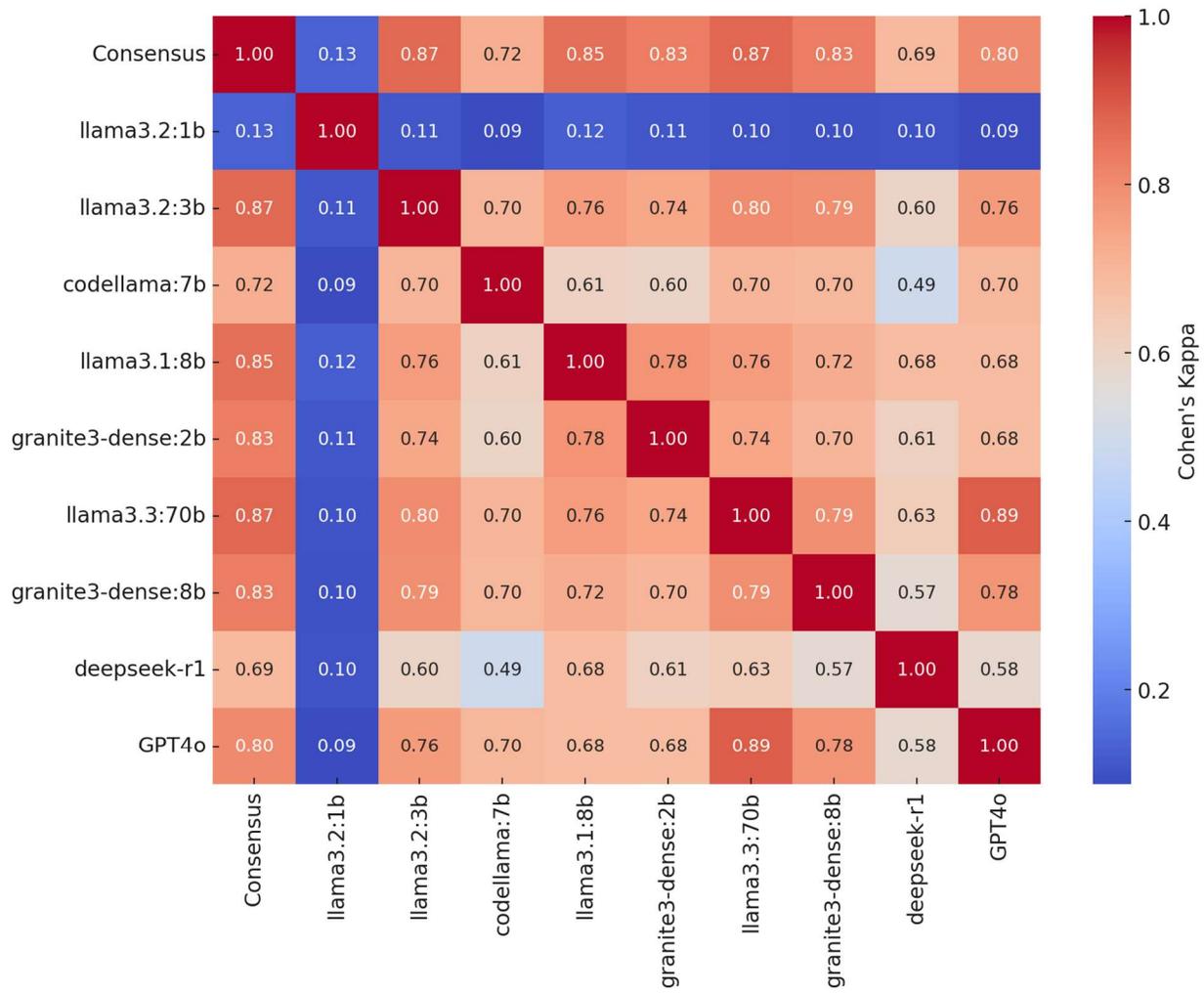

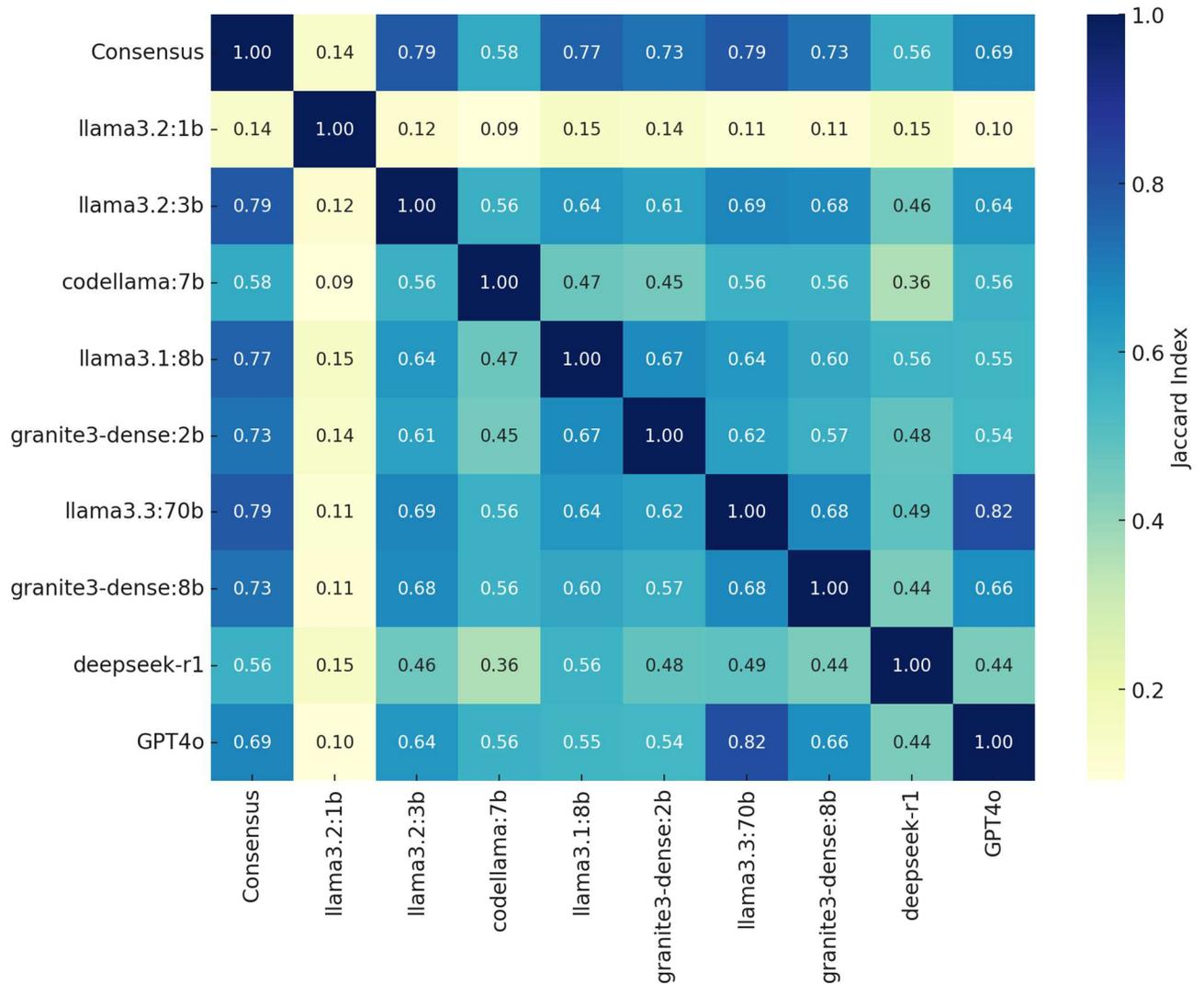

Figure 3. Cohen's kappa (a) and Jaccards similarity index (b) comparing the outputs of the nine large language models, a consensus score (4 or more LLM in agreement) for identifying presence or absence hemorrhage in the impression section of the reports shows agreement ranging from poor to excellent.